\documentclass[runningheads,a4paper]{./LNCS_files/llncs}

\usepackage[french,english]{babel}
\usepackage[utf8]{inputenc}
\usepackage[T1]{fontenc}
\usepackage{amssymb}
\usepackage{graphicx}
\usepackage{amsmath}
\usepackage{epstopdf}
\usepackage{enumitem}
\usepackage{algorithmicx}
\usepackage{algorithm}
\usepackage{algpseudocode}
\usepackage{dsfont}
\usepackage{comment}
\usepackage{url}
\usepackage{cite}
\usepackage{lineno}

\usepackage{graphicx}
\usepackage{subcaption}
\usepackage[utf8]{inputenc}
\usepackage[export]{adjustbox}
\usepackage{wrapfig}

\usepackage[final]{changes}
\definechangesauthor[name={IK}, color=blue]{jb}

\newcommand{\R}{\mathbb{R}}

\newcommand{\V}{\mathcal{V}}
\newcommand{\EE}{\mathcal{E}}

\newcommand{\Na}{\mathbb{N}^{\ast}}

\newcommand{\xb}{\mathbf{x}}
\newcommand{\yb}{\mathbf{y}}

\newcommand{\wb}{\mathbf{w}}
\newcommand{\zb}{\mathbf{z}}

\newcommand{\gammab}{\boldsymbol\gamma}

\newcommand{\thetab}{\boldsymbol\theta}

\newcommand{\Id}{\mathrm{Id}}

\newcommand{\epsilonb}{\boldsymbol\varepsilon}
\newcommand{\simiid}{\mathop{\sim} \limits^{\mathrm{i.i.d.}}}

\newcommand{\G}{\mathcal{G}}
\newcommand{\D}{\mathcal{D}}

\begin{document}

\mainmatter  

\title{Statistical learning of spatiotemporal patterns from longitudinal manifold-valued networks}

\titlerunning{Spatiotemporal patterns learning from manifold-valued networks}

%
%

\author{I. Koval\inst{3,1}, J.-B. Schiratti\inst{3,1}, A. Routier\inst{1}, M. Bacci\inst{1}, O. Colliot\inst{1,2}, S. Allassonni\`{e}re\inst{3}, S. Durrleman\inst{1}, the Alzheimer's Disease Neuroimaging Initiative}


\authorrunning{I. Koval et al}

\institute{Inria Paris-Rocquencourt, Inserm U1127, CNRS UMR 7225, Sorbonne Universit\'{e}s, UPMC Univ Paris 06 UMRS 1127, Institut du Cerveau et de la Moelle \'{e}pini\`{e}re, ICM, F-75013, Paris, France \and AP-HP, Piti\'{e}-Salp\^{e}tri\`{e}re Hospital, Departments of Neurology and Neuroradiology, F-75013, Paris, France \and INSERM UMRS 1138, Centre de Recherche des Cordeliers, Université Paris Descartes, Paris, France}

%
%

\toctitle{TOC title?}
\tocauthor{TOC author?}
\maketitle

\begin{abstract}

We introduce a mixed-effects model to learn spatiotemporal patterns on a network by considering longitudinal measures distributed on a fixed graph. The data come from repeated observations of subjects at different time points which take the form of measurement maps distributed on a graph such as an image or a mesh. The model learns a typical group-average trajectory characterizing the propagation of measurement changes across the graph nodes. The subject-specific trajectories are defined via spatial and temporal transformations of the group-average scenario, thus estimating the variability of spatiotemporal patterns within the group. To estimate population and individual model parameters, we adapted a stochastic version of the Expectation-Maximization algorithm, the MCMC-SAEM. The model is used to describe the propagation of cortical atrophy during the course of Alzheimer’s Disease. Model parameters show the variability of this average pattern of atrophy in terms of trajectories across brain regions, age at disease onset and pace of propagation. We show that the personalization of this model yields accurate prediction of maps of cortical thickness in patients.

\end{abstract}

\section{Introduction}

There is a great need to understand the progression of Alzheimer's Disease (AD) especially before the clinical symptoms to better target therapeutic interventions \cite{hampel2017}. During this silent phase, neuroimaging reveals the disease effects on brain structure and function, such as the atrophy of the cortex due to neuronal loss. However, the precise dynamics of the lesions in the brain are not so clear at the group level and even less at the individual level. Personalized models of lesion propagation would enable to relate structural or metabolic alterations to the clinical signs, offering ways to estimate stage of the disease progression in the pre-symptomatic phase. Numerical models have been introduced to describe the temporal and the spatial evolution of these alterations, defining a \textit{spatiotemporal trajectory} of the disease, \textit{i.e.} a description of the changes in the brain over time, such as lesion progressions, tissue deformation and atrophy propagation. 

Statistical models are well suited to estimate distributions of spatiotemporal patterns of propagation out of series of short-term longitudinal observations \cite{donohue2014, bilgel2016}. However, the absence of time correspondence between patients is a clear obstacle for these types of approaches. Using data series of several individuals requires to re-align the series of observations in a common time-line and to adjust to a standardized pace of progression. Current models either consider a sequential propagation \cite{young2015}, without taking into account the continuous dynamics of changes, or develop average scenarios \cite{guerrero2016, iturria2015}. Recently, a generic approach to align patients has been proposed for a set of biomarkers in \cite{schiratti2015nips}: the temporal inter-subject variability results from individual variations of a common time-line granting each patient a unique age at onset and pace of progression. On top of the time-alignment of the observations, there exists a spatial variability of the signal propagation that characterizes a distribution of trajectories.

In order to exhibit a spatial representation of the alterations, we study medical images or image-derived features taking the form of a signal discretized at the vertices of a mesh, for instance the cortical thickness distributed on the mesh of the pial surface or Standardized Uptake Value Ratio (SUVR) distributed on the regular voxel grid of a PET scan. The spatial distribution of the signal is encoded in a distance matrix, giving the physical distance between the graph nodes. A sensible prior to include in the model is to enforce smooth variations of the temporal profile of signal changes across neighbouring nodes, highlighting a propagation pattern across the network as in \cite{raj2012}. Extending directly the model in \cite{schiratti2015nips} may lead to an explosion of the number of parameters proportional to the mesh resolution. At infinite resolution, the parameters take the form of a smooth continuous map defined on the image domain. In this paper, we propose to constrain these maps to belong to a finite-dimensional Hilbert Space, penalizing high frequency variations. In practice, these maps are generated by the convolution of parameter values at a sparse set of control nodes on the network with a smoothing kernel. The number of control nodes, whose distribution is determined by the bandwidth of the kernel, controls the complexity of the model regardless of the mesh resolution. Furthermore, the propagation of non-normalized signal could not adequately be modeled by the same curve shifted in time as in \cite{schiratti2015nips}. We introduce new parameters to account for smooth changes in the profiles of changes at neighbouring spatial locations. 


We introduce a mixed-effect generative model that learns a distribution of spatiotemporal trajectory from series of repeated observations. The model evaluates individual parameters (time reparametrization and spatial shifts) that enables the reconstruction of individual disease propagation through time. This non-linear problem is tackled by a stochastic version of the EM algorithm, the MCMC-SAEM \cite{kuhn2005, allassonniere2010} in a high-dimensional setting. It considers fixed-effects describing a group-average trajectory and random effects characterizing individual trajectories as adjustment of the mean scenario. It is used to detect the cortical thickness variations in MRI data of MCI converters from the ADNI database.



\label{sec:Model}
\begin{figure}
  \centering
  \includegraphics[height=4cm]{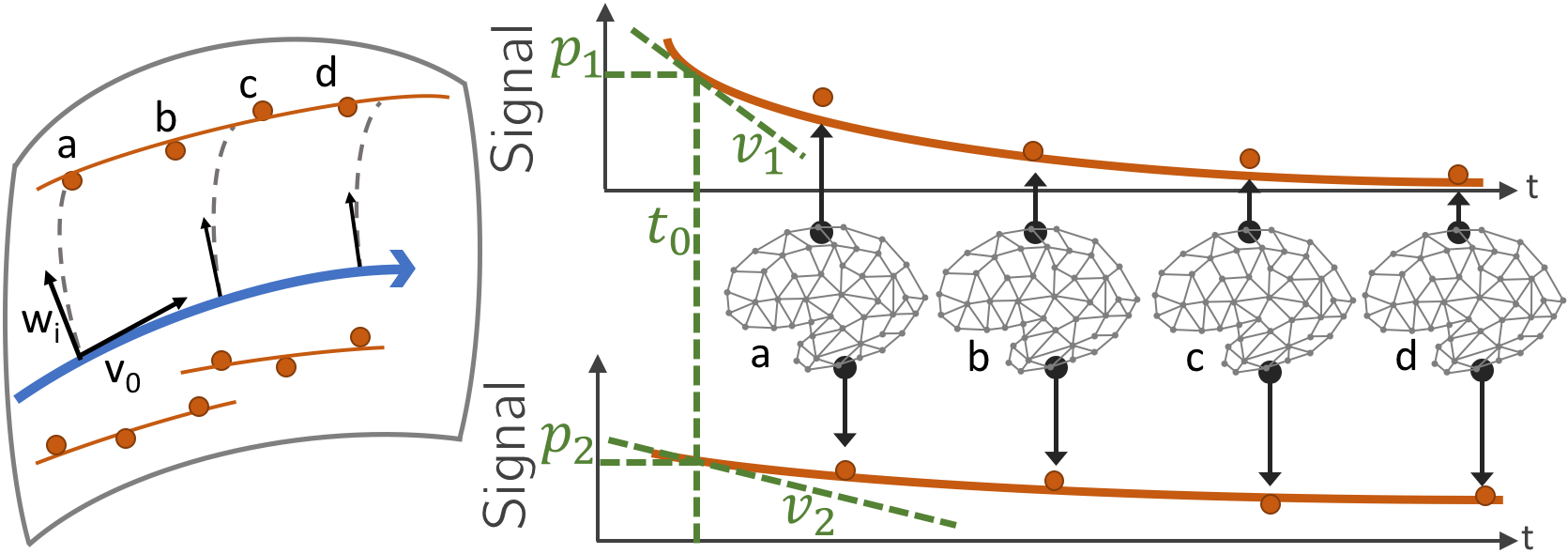}
  \caption{Manifold representation of the mesh observations (left). Orange dots are patient real observations. The blue line is the reconstruction of the mean propagation. The signal value at each node (right), parametrized by ($p, t, v$), allows the reconstruction of the propagation over the network (orange lines)}
  \label{fig:Combinaison}
\end{figure}

\section{Manifold-valued networks}

In the following, we consider a longitudinal dataset $\yb = (\yb_{i,j})_{1 \leq i \leq p, \, 1 \leq j \leq k_i}$of $p$ individuals, such that the $i$th individual is observed at $k_i$ repeated time points $t_{i,1} < \ldots < t_{i,k_i}$. We assume that each observation $\yb_{ij}$ takes the form of $N_v \in \Na$ scalar measures $\big( (\yb_{i,j})_1, ..., (\yb_{i,j})_{N_v} \big)$ referred to as a \textit{measurement map}.

\subsubsection{Manifold-valued measurements distributed on a fixed graph} 
\hfill\\
Let $\mathcal{G} = (\V, \EE)$ be a non-oriented graph where $\V = (\xb_1, ..., \xb_{N_v})$ is a set of vertices of a mesh in $\R^3$ and $\EE$ is a subset of pairs of vertices defining the graph edges. We assume that $\mathcal{G}$  is a common fixed graph such that the $k$th coordinate of each measurement map $y_{ij}$ corresponds to the vertex $\xb_k \in \V$. As the graph corresponds to measurements spatially distributed on a mesh, the edges embed a spatial configuration. Therefore, any edge $(\xb_i, \xb_j)$ is valued with $d$, a geodesic distance on the graph, defining a distance matrix $\D$ such that for all $i,j \in  \{1, .., N_v\}$, $\D_{i,j} = d(\xb_i, \xb_j)$. Each measurement map $\yb_{i,j}$ produces a network $(\G,\D, \yb_{i,j})$, \textit{i.e.} a fixed graph with one-dimensional values associated to each vertex and with distances associated to each edge. 


We consider that the measurements of the patients at each node $k$ corresponds to observations of a signal function $t \mapsto \gammab_k(t)$ at particular time points. Thus, the function $t \mapsto \gammab(t) = (\gammab_1(t), ..., \gammab_{N_v}(t) )$ describes the evolution of the signal over the whole network. We assume that each signal function is continuous and that measurement map $\yb_{i,j} \in \yb$ lies in a space defined by smooth constraints, as expected for bounded or normalized observations (eg. volume ratios, thickness measures, SUVR). Therefore, the space of measurements is best described as a \textit{Riemannian manifold}~\cite{docarmo1992,lee2003}, leading to consider each function $t \mapsto \gammab_k(t)$ as a one-dimensional geodesically complete Riemannian manifold $(M,g^M)$ such that all spatial observation $\yb_{i,j}$ is a point in the product manifold $M^{N_v}$. It follows that for each $i, j$, $(\G,\D, \yb_{i,j})$ is a manifold-valued network.

\subsubsection{Spatial smoothness of the propagation}
\label{section:smoothness}
\hfill\\
Besides the temporal smoothness of the propagation, we expect the signal to be similar for neighbour nodes. We consider that each node is described by $N_p$ parameters that parametrize the signal trajectory. In order to ensure smooth variations of the parameters values at neighbouring nodes, we assume that they result from the interpolation of the parameter values at a sparse subset of uniformly distributed nodes  $\V_C = (\xb_{d_1}, ..., \xb_{d_{N_c}})$, called control nodes. For each parameter $p$, potentially estimated at each node, the control nodes define a parameter evaluation function $p(\xb)$ encoding for all the nodes: $ \forall \xb \in \V$, $\forall \yb \in \V_C$, $\forall i \in \lbrace 1,\ldots, N_c \rbrace$ $p(\xb) = \sum_{i=1}^{N_C} K(\xb, \xb_{d_i}) \beta_i$ and $p(\yb) = p_{\yb}$ where the $(\beta_i)_{1 \leq i \leq N_c}$ are the new model parameters and $K$ is a Gaussian Kernel such that $\forall \xb, \yb \in \V K(\xb, \yb) = f(\frac{d(\xb, \yb)}{\sigma})$, $d$ being the geodesic distance on the graph and $\sigma$ the kernel bandwidth. 

This convolution guarantees the spatial regularity of the signal propagation. Moreover this smooth spatial constraint enables a reduction of the number of parameters, reducing the dimensional complexity from $N_p$ independent parameters at each node, to $N_p$ parameters only at the control nodes.

\section{The statistical model}


\subsubsection{A propagation model} \label{ssec:prop_model}
\hfill\\
Given a set of manifold-valued networks $(\G,\D, \yb)$, the model describes a group-average trajectory in the space of measurements, defined by a geodesic $\gammab$ that allows to estimate a typical scenario of progression. Individual trajectories derive from the group-average scenario through spatiotemporal transformations: the \textit{exp-parallelization} and the \textit{time reparametrization}. 

First, to describe disease pace and onset specific to each subject, we introduced a temporal transformation, called the time-warp, that is defined, for the subject $i$, by $\psi_i(t) = \alpha_i(t_{i,j} - \tau_i - t_0) + t_0$ where $t_0$ is the reference time-point in the space of measurements. The parameter $\tau_i$ corresponds to the time-shift between the mean and the individual age at onset and  $\alpha_i$ is the acceleration factor that describes the pace of an individual, being faster or slower than the average. This \textit{time reparametrization} allows to reposition the dynamics of the average scenario in the real time-line of the $i$th individual.

The \textit{exp-parallelization} allows to translation the observations in the space of measurements, from the mean scenario to individual trajectories, encoding a variation in the trajectory of changes across the nodes of the graph. This exp-parallelization is handled by a family of individual vectors $(\wb_i)_{1 \leq i \leq p}$, called space-shifts. As shown on Figure \ref{fig:Combinaison} (left), the orange dots refer to individual observations in the space of measurements. The group-average trajectory estimated from the longitudinal measurements corresponds to the blue line. The space shifts characterize a spatial shift perpendicular to $v_0$ that describes the velocity of the mean scenario. 

Finally, the parameters $(\alpha_i, \tau_i, \wb_i)$ allow the reconstruction of the individual trajectories from the mean scenario of propagation. 

Given a noise $\epsilonb_{i,j} \simiid \mathcal{N}(\mathbf{0}, \sigma^2 \Id_{N_v})$, the mixed-effect model writes, for a arbitrary vertex function $\gammab_k(t)$:

\begin{equation}
(\yb_{i,j})_{k} = \gamma_k\Bigg( \frac{(\wb_i)_k}{\dot{\gamma_k}(t_0)}  + \alpha_i(t_{i,j} - t_0 - \tau_i) + t_0 \Bigg) + (\epsilonb_{i,j})_k
\label{eq:propagationmodel}
\end{equation}


\subsubsection{Parameters estimation with the MCMC-SAEM algorithm}
\label{sec:MCMC-SAEM}
\hfill\\
To reconstruct the long-term scenario of the disease propagation, we estimate the parameters of the group-average trajectory $\thetab = ( (\beta_{param}^i)_{1 \leq i \leq N_c , 1 \leq j \leq N_p}, \sigma )$ using a maximum likelihood estimator. The random-effects $\zb = (z_i)_{1 \leq i \leq p} = (\wb_i, \alpha_i, \tau_i)_{1 \leq i \leq p}$ are considered as latent variables, whose distributions characterize the variability of the individual trajectories. Due to the non-linearity in Eq. ~\eqref{eq:propagationmodel}, we use a Stochastic Approximation Expectation Maximization \cite{delyon1999} coupled with a Monte-Carlo Markov Chain sampler (MCMC-SAEM) \cite{kuhn2004}. Let $\theta^{(k)}$ be the current estimation of the parameters and $\zb^{(k)}$ the current iterate of the Markov chain of the latent variables. The algorithm alternates between a simulation step, a stochastic approximation step and a maximization step, until convergence \cite{allassonniere2010}. The simulation uses an adaptive version \cite{atchade2006} of the Hasting Metropolis within Gibbs sampler to draw $\zb^{(k+1)}$ from $(\zb^{(k)}, \yb, \thetab^{(k)})$. This algorithm was chosen as it handles non-linear mixed effects models \cite{kuhn2005} with proven convergence and consistent estimations in practice.

\begin{figure}
  \centering
  \includegraphics[height=8cm]{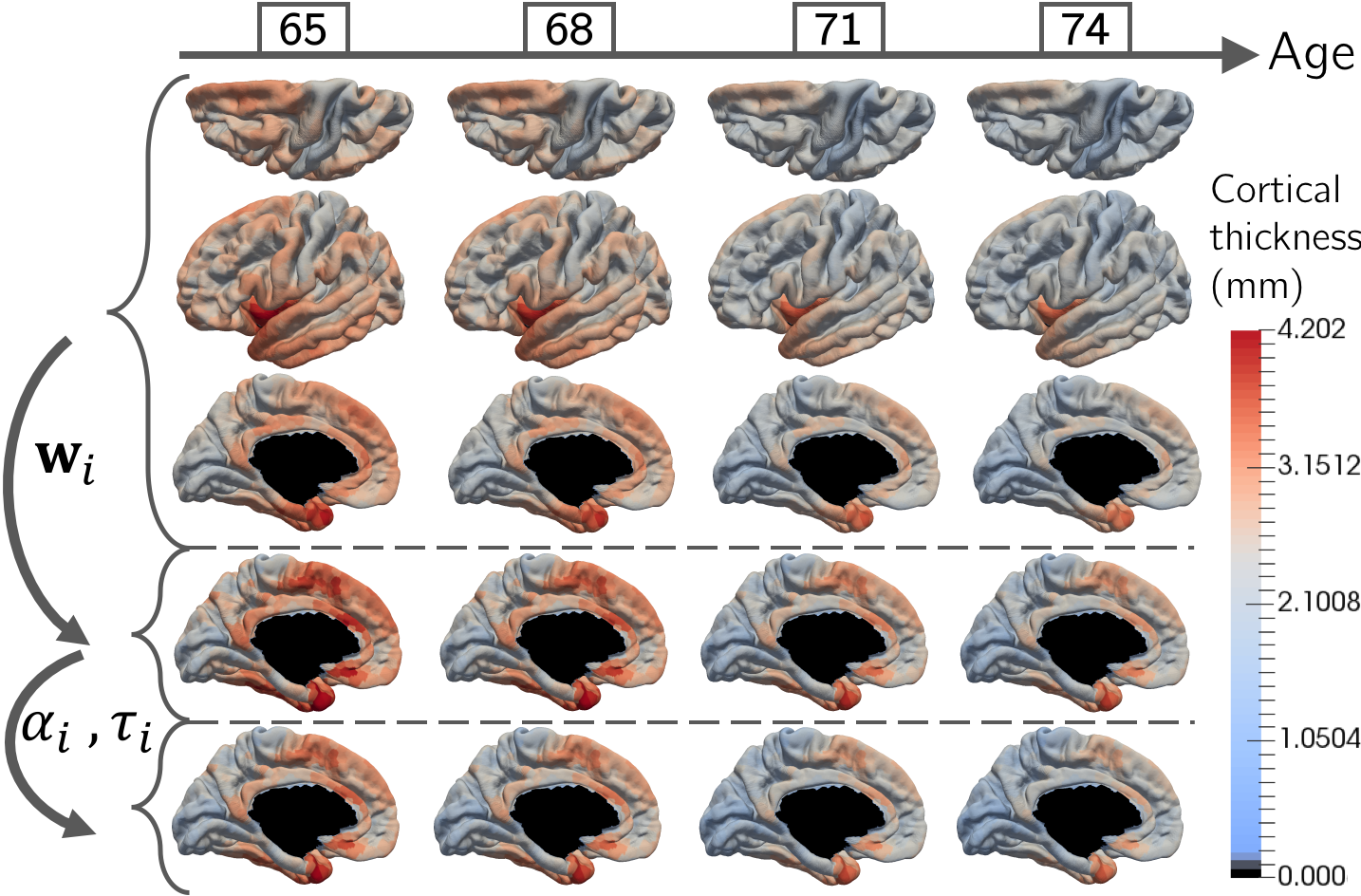}
  \caption{Cortical thickness at 65, 68, 71 and  74 years old of the mean propagation (first rows). Effect of the space-shift $\wb_i$ (fourth row), then with temporal reparametrization $\alpha_i, \tau_i$ (fifth row) on the cortical thickness.}
  \label{fig:MeanProgression}
\end{figure}

\subsubsection{Model instantiation}
\label{Instantiation}
\hfill\\
As many measurements correspond to positive values (eg. the cortical thickness, volume ratios), we consider in the following the open interval $M = ]0,+\infty[$ as a one-dimensional Riemannian manifold equipped with a Riemannian metric $g$ such that for all $ p \in M $ and for all $(u,v) \in \mathrm{T}_{p}M$, $g_p(u,v) = uv/p^2$. With this metric and given $k \in \lbrace 1,\ldots, N_v \rbrace$, $\mathrm{M}$ is a geodesically complete Riemannian manifold whose geodesics are of the form  $t \mapsto \gammab_k(t) = p_k \exp(\frac{v_k}{p_k}(t - t_k) ) \,$ where $p_k \in M$, $t_k \in \R, v_k \in \mathrm{T}_{p_k}M$. These parameters are represented on Figure \ref{fig:Combinaison} (right) at two nodes where the decrease of the signal varies spatially. For identifiability reasons, we choose to fix the parameters $t_k$ among the nodes, leading to a shared parameter $t'_0$ such that for all $k \in \lbrace 1,\ldots, N_v \rbrace$ $t_k = t'_0$. As $t'_0$ can be arbitrarily chosen in $\R$, we fix $t'_0 = t_0$ defined in Section \ref{ssec:prop_model}. Considering the interpolation functions introduced in Section \ref{section:smoothness} and the fact that the parameters $(p^k_j)$ are $(p_k, v_k)$, it leads to define $p(\xb) = \sum_{i=1}^{N_C} K(\xb, \xb_{d_i}) \beta_{p}^i$ and $v(\xb) = \sum_{i=1}^{N_C} K(\xb, \xb_{d_i}) \beta^i_{v}$

Finally, the model defined in ~\eqref{eq:propagationmodel} rewrites:
\begin{equation}
(\yb_{i,j})_k = p(\xb_k) \exp\Bigg( \frac{(\wb_i)_k}{p(\xb_k)} + \frac{v(\xb_k)}{p(\xb_k)}\alpha_i(t_{i,j} - t_0 - \tau_i)\Bigg) + (\epsilonb_{i,j})_k
\label{eq:propagation model exp prop}
\end{equation}

such that $\thetab = (t_0, (\beta^i_{p})_{1 \leq i \leq N_c}, (\beta^i_{v})_{1 \leq i \leq N_c}, \sigma)$ and $\zb = (\wb_i, \alpha_i, \tau_i)_{1 \leq i \leq p}$


\section{Experimental results}
\label{sec:Experiments}


\subsubsection{Data}
\label{sec:Experiments>Data}
\hfill\\
We used this model to highlight typical spatiotemporal patterns of cortical atrophy during the course of Alzheimer's Disease from longitudinal MRI of 154 MCI converters from the ADNI database, which amounts for 787 observations, each subject being observed 5 times on average. We aligned the measures on a common atlas with FreeSurfer \cite{reuter2012} so that the measurement maps are distributed on the same common fixed-graph $\G$ which is constituted of 1827 nodes that map the surface of the left pial surface. Out of the vertices, we selected 258 control nodes uniformly distributed over the surface. They encode the spatial interpolation of the propagation. The distance matrix $\D$ is defined by a geodesic distance on $\G$.


\begin{figure}
\begin{subfigure}{0.5\textwidth}
\includegraphics[width=0.90\linewidth]{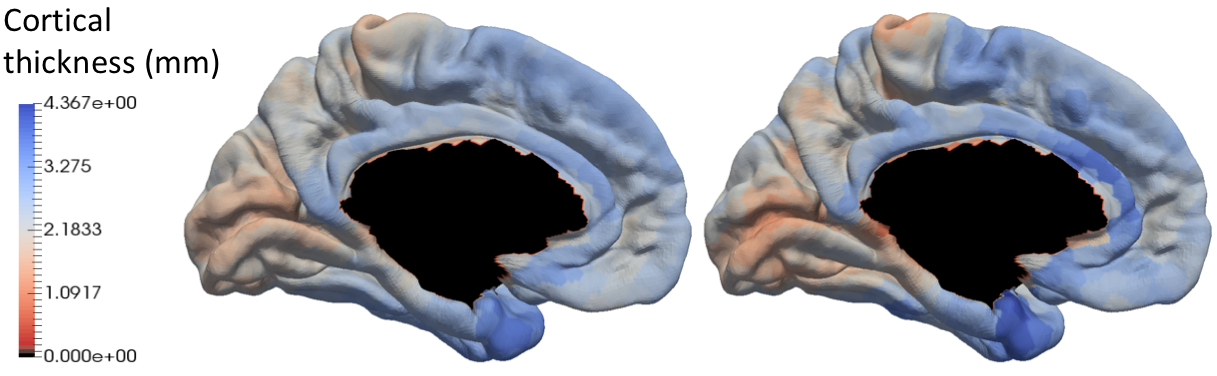} 
\caption{Real observation of the cortical thickness (right) and its model reconstruction (left)}
\label{fig:subim1}
\end{subfigure}
\begin{subfigure}{0.5\textwidth}
\includegraphics[width=0.90\linewidth]{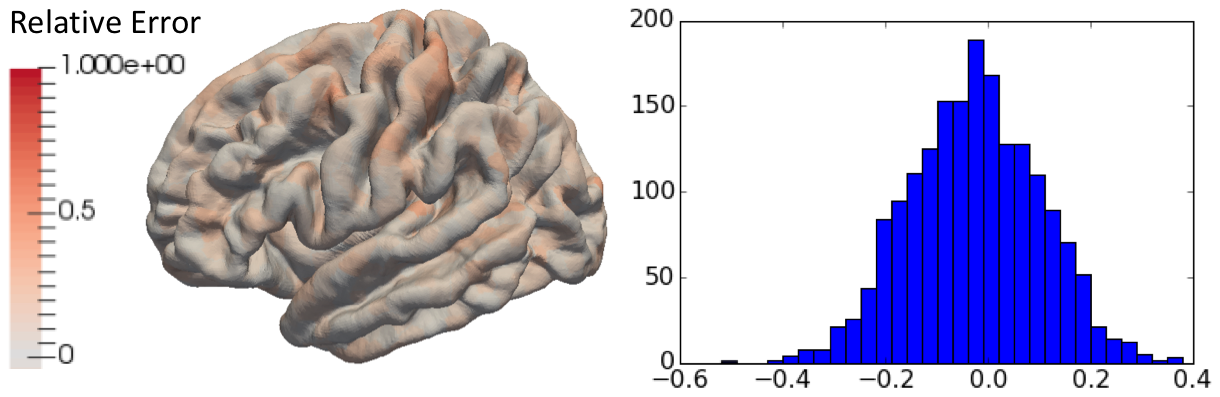}
\caption{Relative error between the observation and its reconstruction plotted on the mesh (left) and its histogram (right)}
\label{fig:subim2}
\end{subfigure} 
\caption{Comparison of an observation and its reconstruction by the model}
\label{fig:results1}
\end{figure}

\subsubsection{Cortical thickness measurements}
\label{sec:Experiments>CortThick}

 

We used the model instantiation defined in Section \ref{Instantiation} to characterize the cortical thickness decrease. Multiple runs of 30.000 iterations ($\sim$4hours) of this MCMC-SAEM lead to a noise standard deviation $\sigma \simeq 0.27$mm with $90\%$ of the data included in $[1.5,3.6]$ mm. The mean spatiotemporal propagation, described on the first three rows of the Figure \ref{fig:MeanProgression} as the cortical thickness at respectively 65, 68, 71 and 74 years old shows that the primarily affected area is the medial-temporal lobe, followed by the temporal neocortex. The parietal association cortex and the frontal lobe are also subject to important alterations. On the other side, the sensory-motor cortex and the visual cortex are less involved in the lesion propagation. These results are consistent with previous knowledge of the Alzheimer's Disease effects on the brain structure.
As the model is able to exhibit individual spatiotemporal patterns with their associated pace of progression, the fourth and fifth rows of the Figure \ref{fig:MeanProgression} represent consecutively the effect of the parallel shifting and of the time reparametrization on the cortical thickness atrophy. The figure \ref{fig:subim1} shows the real cortical thickness of a subject and the reconstruction predicted by the model. The relative error and its histogram are represented on Figure \ref{fig:subim2}. The reconstruction is coherent to the real observation, the remaining error represents essentially an unstructured noise that we precisely try to smooth out.

  
\section{Discussion and perspectives}
\label{sec:Discussion}
 
We proposed a mixed-effect model which is able to evaluate a group-average spatiotemporal propagation of a signal at the nodes of a mesh thanks to longitudinal neuroimaging data distributed on a common network. The network vertices describe the evolution of the signal whereas its edges encode a distance between the nodes via a distance matrix. The high dimensionality of the problem is tackled by the introduction of control nodes: they allow to evaluate a smaller number of parameters while ensuring the smoothness of the signal propagation through neighbour nodes. Moreover, individual parameters characterize personalized patterns of propagation as variations of the mean scenario. The evaluation of this non-linear high dimensional model is made with the MCMC-SAEM algorithm that leads to convincing results as we were able to highlight areas affected by considerable neuronal loss: the medial-temporal lobe or the temporal neocortex. 

The distance matrix, which encodes here the geodesic distance on the cortical mesh, may be changed to account for the structural or functional connectivity information. In this case, signal changes may propagate not only across neighbouring locations, but also at nodes far apart in space but close to each other in the connectome. The model can be used with multimodal data, such as PET scans, introducing numerical models of neurodegenerative diseases that could inform about the disease evolution at a population level while being customizable to fit individual data, predicting stage of the disease or time to symptom onset. 

\textbf{This work has been partly funded by ERC grant N\textsuperscript{o}678304, H2020 EU grant N\textsuperscript{o}666992, and ANR grant ANR-10-IAIHU-06.}


\bibliographystyle{LNCS_files/splncs03}

\bibliography{biblio}

\end{document}